\documentclass{article}
\usepackage{ijcai23}

\usepackage{times}
\usepackage{soul}
\usepackage{url}
\usepackage[hidelinks, colorlinks, linkcolor=blue, citecolor=black]{hyperref}
\usepackage[utf8]{inputenc}
\usepackage[small]{caption}
\usepackage{graphicx}
\usepackage{amsmath}
\usepackage{amsthm}
\usepackage{booktabs}
\usepackage{algorithm}
\usepackage{algorithmic}
\usepackage[switch]{lineno}
\usepackage{booktabs}
\usepackage{adjustbox}

\usepackage{color}
\usepackage{xcolor}
\usepackage{tikz}
\usepackage[edges]{forest}
\definecolor{hidden-draw}{RGB}{20,68,106}
\definecolor{hidden-pink}{RGB}{255,245,247}
\definecolor{green}{HTML}{2a9d8f}

\definecolor{object_color}{HTML}{00CC00}
\definecolor{attribute_color}{HTML}{FF0080}
\definecolor{relation_color}{HTML}{007FFF}

\title{A Survey on Hallucination in Large Vision-Language Models}

\author{
    Hanchao Liu\footnote{Equal Contribution}\and Wenyuan Xue$^*$\and Yifei Chen\and Dapeng Chen\and Xiutian Zhao\and \\Ke Wang\and Liping Hou\and Rongjun Li\and Wei Peng
    \affiliations
    IT Innovation and Research Center, Huawei Technologies
    \emails
    \{liuhanchao2, xuewenyuan1, chenyifei14, chendapeng8, zhaoxiutian, \\wangke215, houliping1, lirongjun3, peng.wei1\}@huawei.com
}

\begin{document}

\maketitle

\begin{abstract}
Recent development of Large Vision-Language Models (LVLMs) has attracted growing attention within the AI landscape for its practical implementation potential. However, ``hallucination'', or more specifically, the misalignment between factual visual content and corresponding textual generation, poses a significant challenge of utilizing LVLMs. In this comprehensive survey, we dissect LVLM-related hallucinations in an attempt to establish an overview and facilitate future mitigation. Our scrutiny starts with a clarification of the concept of hallucinations in LVLMs, presenting a variety of hallucination symptoms and highlighting the unique challenges inherent in LVLM hallucinations. Subsequently, we outline the benchmarks and methodologies tailored specifically for evaluating hallucinations unique to LVLMs. Additionally, we delve into an investigation of the root causes of these hallucinations, encompassing insights from the training data and model components. We also critically review existing methods for mitigating hallucinations. The open questions and future directions pertaining to hallucinations within LVLMs are discussed to conclude this survey. To keep track of this field and continuously update our survey, we maintain a repository of relevant references at \url{https://github.com/lhanchao777/LVLM-Hallucinations-Survey}.
\end{abstract}

\section{Introduction}
In the rapidly evolving realm of artificial intelligence, large language models (LLMs) such as GPT-4~\cite{GPT-4}, LLaMA~\cite{LLaMA}, and LLaMA2~\cite{touvron2023llama} have made remarkable strides in natural language understanding (NLU) and generation (NLG). To harness the NLU and NLG capabilities of LLMs for vision-language tasks, a prevalent approach is inserting visual features as supplementary inputs to LLMs and aligning them with textual features. This method has been adapted in several large vision-language models (LVLMs) such as MiniGPT-4~\cite{MiniGPT-4}, LLaVA~\cite{LLaVA}, and LLaVA-1.5~\cite{LLaVA-1.5}. Despite the promising results demonstrated by existing LVLMs, an unignorable issue has been impeding their practical application: \textit{hallucination}. Hallucination in LVLM denotes a disagreement between factual content of images and corresponding generated textual content, akin to the solely textual hallucination encountered in large language models~\cite{huang2023survey}.

\begin{figure}
    \centering
    \includegraphics[width=\linewidth]{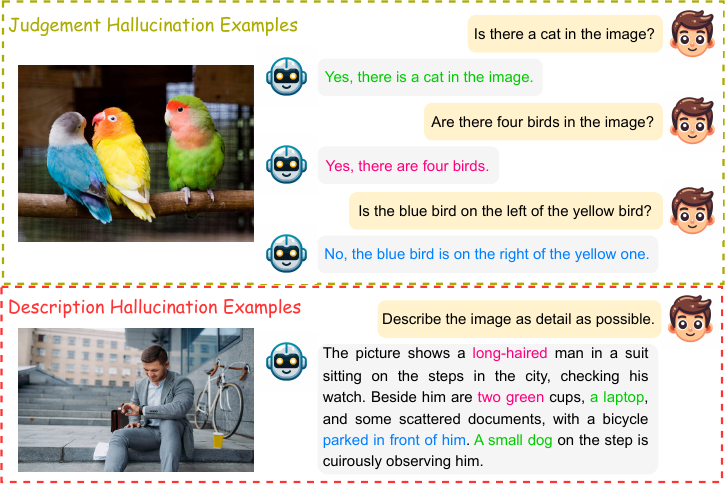}
    \caption{Hallucination examples in LVLMs. Hallucination symptoms may manifest as deficiencies in various vision-language tasks like \textit{judgment} and \textit{description}, or factual errors in different visual semantics, such as  \textcolor{object_color}{objects}, \textcolor{attribute_color}{attributes}, and \textcolor{relation_color}{relations}.}
    \label{fig:hallucination_types}
\end{figure}
 
Existing studies \cite{CHAIR,POPE,CIEM,HallE-Switch} have addressed the issue of hallucination in image captioning models with a predominant focus on the ``presence of objects'', specifically whether the objects depicted in a given image are accurately described by the text generated by the model. In contrast to image captioning models trained within closed domains, LVLMs leverage the robust understanding and expressive capabilities of LLMs to obtain more detailed and fluent generated descriptions. However, these enhanced abilities also diversify and potentially exacerbate hallucination, which is not only limited to object existence, but also manifested in the descriptive errors such as attribute and relation errors. We focus on visual hallucinations, referring to all inconsistencies between the semantic content conveyed by the image and the textual content generated by the model.

Hallucination symptoms in LVLMs are multifaceted. From a cognitive perspective, hallucinations can manifest as errors in true/false judgments and inaccuracies in the description of visual information. For instance, as illustrated in the first example of Figure~\ref{fig:hallucination_types}, the model ill responds to questions such as ``Is there a cat in the image?" and ``Are there four birds in the image?", demonstrating flawed factual discernment. Additionally, the second example shows an misalignment of  the generated description and the visual facts. Meanwhile, a visual semantics perspective provides us with a ternary taxonomy: hallucination on object, attribute, and relationship. For example, the model generates non-existent objects in the image such as ``laptop'' and ``small dog'', provides incorrect attribute descriptions such as describing the man being ``long-haired'', and makes inaccurate assertions about the relationships between objects such as claiming the bicycle is ``in front of'' the man. Current methods evaluate these hallucination in LVLMs based on the models' cognitive performance, focusing on two main aspect: non-hallucinatory generation and hallucination discrimination. The former involves a detailed analysis of the hallucinatory elements in the model's response and quantifying their proportion. The latter, on the other hand, only requires a binary judgment of whether the response comprises any hallucinatory content. These approaches are comprehensively discussed in \S \ref{sec:eval}.

While the causes of hallucinations in LLMs~\cite{huang2023survey} have been discussed extensively, LVLMs' visual modality introduces unique challenges in analyzing these occurrences. We conduct a thorough analysis of hallucinations in LVLMs, focusing on the training data and model characteristics. Our analysis indicates that hallucinations in LVLMs are caused not only by the generative nature of LLMs but also by biased training data, the inability of vision encoders to accurately ground images, misalignment among different modalities, insufficient context attention, and many other factors. Following this, we provide a comprehensive overview of the existing hallucination mitigation methods. Corresponding to the causes, current mitigation approaches predominantly focus on the optimization of training data, refinement of various modules within LVLMs, and post-processing of generated outputs. These methods are employed to diminish the occurrence of hallucinations, thereby yielding more faithful responses. In the end, we list several considerable directions for developing the research of hallucinations in LVLMs.

In summary, this study aims to provide insights for the development of LVLMs and explore the opportunities and challenges related to LVLM hallucinations. This exploration not only helps us understand the limitations of current LVLMs, but also offers important guidance for future research and the development of more reliable and efficient LVLMs.

\section{Hallucination in the Era of LVLM}

\subsection{Large Vision-Language Models}
\label{sec:lvlms}

LVLMs are advanced multimodal models that process visual and textual data to solve compound tasks involving both vision and natural language. Incorporating LLMs' capabilities, LVLMs are an evolution of previous Vision-Language Pre-trained Models (VLPMs)~\cite{long2022vision}.

LVLM architectures typically comprise three components: a visual encoder, a modality connection module, and a LLM. The visual encoder, often an adaptation of the CLIP vision encoder~\cite{CLIP}, transforms input images into visual tokens. The connection module is designed to align visual tokens with the word embedding space of the LLM, ensuring that the LLM can handle the visual information. There are various methods for modality alignment, including cross-attention~\cite{Flamingo}, adapters~\cite{gao2023llama}, Q-Formers~\cite{BLIP2,InstructBLIP,MiniGPT-4}, and simpler structures such as linear layers or multi-layer perceptrons (MLP)~\cite{LLaVA,chen2023minigpt,LLaVA-1.5}. The LLM functions like the central processing unit within LVLMs, receiving aligned visual and textual information and subsequently synthesizing this information to produce responses.

The training of LVLMs involves two key stages: (1) pre-training, where LVLMs acquire vision-language knowledge from aligned image-text pairs, and (2) instruction-tuning, during which LVLMs learn to follow human instructions using a varied task dataset. Following these stages, LVLMs can efficiently process and interpret visual and textual data, enabling them to reason about concepts in compound multimodal tasks like Visual Question Answering (VQA).

\subsection{Hallucination in LVLMs}
\label{sec:hallucination_lvlms}

Hallucination in LVLMs refer to contradictions between the visual input (taken as `fact') and the textual output of a LVLM. Through the lens of visual-language tasks, LVLM hallucination symptoms can be interpreted as deficiencies in judgement or description.

A judgement hallucination occurs when the model's response to a user's query or statement is in disagreement with the actual visual data. For example, as shown in Figure~\ref{fig:hallucination_types}, when being presented with an image depicting three birds and inquired whether there is a cat in the picture, the model wrongly affirms with a ``Yes''. A description hallucination, on the other hand, is a failure to faithfully depict the visual information. As exemplified in lower part of Figure~\ref{fig:hallucination_types}, where the model inaccurately describes the man’s hair, the quantity and color of cups, the bicycle's location, and fabricates non-existent objects such as laptop and dog. 

From a semantic perspective, such misalignment can be characterized by claiming nonexistent objects, incorrect object attributes, or inaccurate object relations, as highlighted in different colors.

\subsection{Unique Challenges regarding Hallucination in LVLMs}
\label{sec:hallucination_comparision} 
LVLMs handle vision-language tasks by incorporating both visual and language modules. However, the integration also pose unique challenges in hallucination detection, causal inference, and mitigation methods.

\paragraph{Hallucination Detection Difficulties} The multimodal nature of LVLM impedes detecting hallucinations. LVLM Hallucinations may manifest across various semantic dimensions, including, but not limited to the object, attribute, and relation~\cite{HallE-Switch,you2023ferret}. To detect these hallucinations comprehensively, the model is tasked not only on natural language understanding but also employing fine-grained visual annotations and aligning them precisely with the generated text.

\paragraph{Intertwined Causes} The causes of hallucinations in LVLMs are often multifaceted. On one hand, there are data-related issues shared by both LLMs and LVLMs~\cite{CIEM} such as misinformation, biases, and knowledge boundary limits. However, LVLMs uniquely suffer from their incorporation of visual data. For instance, visual uncertainty, such as unclear or distorted images, can exacerbate the language priors and statistical biases in LVLMs, leading to severer hallucinations~\cite{liu2023mitigating}.

\paragraph{Compound Mitigation Methods} Besides adapting LLM-targeted hallucination mitigation methods, such as data quality enhancement, encoding optimization, and aligning with human preferences, LVLM-specific methods also include refining visual representations and improving the multi-modal alignment. For instance, enlarging visual resolution has been suggested to effectively reduce hallucinations~\cite{bai2023qwen}. Nonetheless, training high-resolution visual encoders with extensive data can be highly resource-demanding. Therefore, it is imperative to investigate more cost-effective strategies for augmenting visual representations. Moreover, the significant gap between visual and textual tokens suggest that improving the vision-language token alignment can potentially lower the incidence of hallucinations~\cite{jiang2023hallucination}.

\section{Evaluation Methods and Benchmarks}
\label{sec:eval}

\tikzstyle{my-box}=[
    rectangle,
    draw=hidden-draw,
    rounded corners,
    text opacity=1,
    minimum height=1.5em,
    minimum width=5em,
    inner sep=2pt,
    align=center,
    fill opacity=.5,
    line width=0.8pt,
]
\tikzstyle{leaf}=[my-box, minimum height=1.5em,
    fill=hidden-pink!80, text=black, align=left,font=\normalsize,
    inner xsep=2pt,
    inner ysep=4pt,
    line width=0.8pt,
]

\begin{figure*}[!th]
    \centering
    \resizebox{0.96\textwidth}{!}{
        \begin{forest}
            forked edges,
            for tree={
                grow=east,
                reversed=true,
                anchor=base west,
                parent anchor=east,
                child anchor=west,
                base=left,
                font=\large,
                rectangle,
                draw=hidden-draw,
                rounded corners,
                align=left,
                minimum width=4em,
                edge+={darkgray, line width=1pt},
                s sep=3pt,
                inner xsep=2pt,
                inner ysep=3pt,
                line width=0.8pt,
                ver/.style={rotate=90, child anchor=north, parent anchor=south, anchor=center},
            },
            where level=1{text width=13em,font=\normalsize,}{},
            where level=2{text width=9em,font=\normalsize,}{},
            where level=3{text width=12em,font=\normalsize,}{},
            where level=4{text width=10em,font=\smallsize,}{},
            [
                LVLM Hallucination \\ 
                Evaluation Methods, ver
                [
                    Evaluation on \\ Non-Hallucinatory Generation
                    [
                        Handcrafted \\ Pipeline Methods 
                        [
                            {CHAIR}~\cite{CHAIR}{, }\\
                            {CCEval}~\cite{HallE-Switch}{, }\\
                            {FAITHSCORE}~\cite{FAITHScore}
                            , leaf, text width=15em
                        ]
                    ]
                    [
                        Model-based \\ End-to-End Methods
                        [
                            LLM-based Evaluation
                            [
                                {GAVIE}~\cite{liu2023mitigating}{, }\\
                                {MMHal-Bench}~\cite{sun2023aligning}
                                , leaf, text width=15em
                            ]
                        ]
                        [
                            Hallucination Data \\
                            Driven Model Evaluation
                            [
                                {M-HalDetect}~\cite{gunjal2023detecting}{, }\\
                                {HaELM}~\cite{wang2023evaluation}
                                , leaf, text width=15em
                            ]
                        ]
                    ]
                ]
                [
                    Evaluation on \\ Hallucination Discrimination
                    [
                        {POPE}~\cite{POPE}{, }\\
                        {CIEM}~\cite{CIEM}{, }\\
                        {NOPE}~\cite{NOPE}
                        , leaf, text width=15em
                    ]
                ]
            ]
        \end{forest}
    }
    \caption{Taxonomy of Hallucination Evaluation Methods for LVLMs}
    \label{fig:taxo_of_evaluation}
\end{figure*}
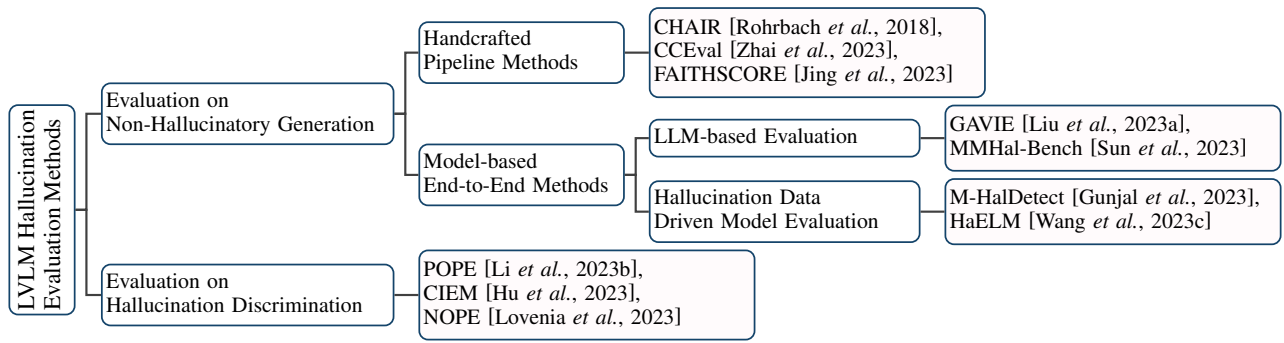

{
\tiny
\begin{table*}
\centering
\caption{Hallucination Evaluation Benchmarks for LVLMs. `Dis' and `Gen' stand for `Discriminative' and `Generative', respectively.}
\label{tab:evaluation_benchmarks}
\resizebox{0.9\textwidth}{!}{%
\begin{tabular}{ccccc}
\toprule
Benchmark & Evaluation & Size & Hallucination Symptoms & Metrics\\
\midrule
POPE \cite{POPE}  & Dis   & 3000 & Object & Accuracy\\
NOPE \cite{NOPE} & Dis & 17983 & Object & Accuracy\\
CIEM \cite{CIEM} & Dis & 72941 & Object & Accuracy\\
M-HalDetect \cite{gunjal2023detecting}  & Gen   & 4000 & Object \& Attribute \& Relation & Reward Model Score\\
GAVIE \cite{liu2023mitigating}  & Gen   & 1000 & Object \& Attribute & Accuracy \& Relevancy\\
FAITHScore \cite{FAITHScore} & Gen & 2000 & Object \& Attribute \& Relation & FAITHScore\\
HaELM \cite{wang2023evaluation} & Gen & 5000 & Not Explicitly Stated & Accuracy\\
MMHal-Bench \cite{sun2023aligning} & Gen & 96 & Object \& Attribute \& Relation & Rating Score\\
AMBER~\cite{wang2023llm} & Dis \& Gen & 15202 & Object \& Attribute \& Relation & AMBER Score\\
\bottomrule
\end{tabular}
}
\end{table*}
}

Establishing the concepts of hallucination in LVLM, we then turn to examine existing LVLM hallucination evaluation methods and benchmarks. 
Corresponding to the hallucination symptoms in description and judgement tasks mentioned in Figure~\ref{fig:hallucination_types}, current evaluation methods can be classified into two major approaches: (1) assessing the model's ability of non-hallucinatory content generation, and (2) evaluating the model's ability of hallucination discrimination, as shown in Figure \ref{fig:taxo_of_evaluation}. Similarly, the benchmarks can also be categorized into discriminative and generative ones on the basis of evaluation tasks, as demonstrated in Table~\ref{tab:evaluation_benchmarks}.

\subsection{Evaluation on Non-Hallucinatory Generation}
\label{sec:evaluation_methods}

Evaluating non-hallucinatory generation is to measure the proportion of hallucinated content in the outputs. Currently, there are primarily two main types: \textit{handcrafted pipeline} and \textit{model-based end-to-end} methods.

\paragraph{Handcrafted Pipeline Methods} Handcrafted pipeline methods feature strong interpretability by manually designing multiple steps with specific and clear objectives. CHAIR~\cite{CHAIR} targets evaluating object hallucinations in image captioning by quantifying differences of objects between model generation and ground-truth captions. However, such conventional methods struggle with LVLMs' vast object categories~\cite{lu2018neural}. To address this, CCEval~\cite{HallE-Switch} employs object alignment module powered by GPT-4 before applying CHAIR. FAITHSCORE \cite{FAITHScore} offers a reference-free and fine-grained evaluation method for evaluating the faithfulness of free-form model responses, which comprises identifying descriptive sub-sentences, extracting atomic facts, and comparing extracted facts against the corresponding input image. 

\paragraph{Model-based End-to-End Methods} End-to-end methods directly assess the performance of LVLMs by evaluating their responses, such as scoring the responses or determining the presence of hallucinations in responses. Existing end-to-end evaluation methods can broadly be categorized into two types. 
The first type is \textit{LLM-based evaluation} which implements an advanced LLM (e.g., GPT-4) to rate LVLM generated content based on hallucination~\cite{liu2023mitigating,sun2023aligning}. These methods leverage the LLM's robust natural language understanding and processing capabilities. By integrating visual information (such as dense captions and object bounding boxes), user instructions, and model responses as input, LLMs are prompted to evaluate and score these responses.
The second type is \textit{hallucination data driven model evaluation} which constructs labelled hallucination datasets for fine-tuning models to detect hallucinations. \cite{gunjal2023detecting} creates the M-HalDetect dataset with annotated LVLM image descriptions and fine-tune an InstructBLIP~\cite{InstructBLIP} model on the dataset for hallucination identification. Similarly, \cite{wang2023evaluation} compiles a dataset of hallucinatory responses generated by ChatGPT and then fine-tune a LLaMA model with LoRA \cite{LoRA} on this dataset to differentiate hallucinatory responses from non-hallucinatory ones.

\subsection{Evaluation on Hallucination Discrimination}
Hallucination discrimination evaluation approach aim to assess the hallucination discrimination ability of LVLMs. The methods follow this approach typically adopt a question-answering format, posing inquiries to LVLMs consisting of descriptions that agree or conflict with the content of the provided input image and assessing the models' responses~\cite{POPE,CIEM,NOPE}. POPE~\cite{POPE} designs binary (\textit{Yes-or-No}) questions about object presence in images such as ``\textit{Is there a person in the image?}'' to evaluate the hallucination discrimination ability of LVLMs. The objects asked in questions are selected under three distinct sampling strategies: random (selecting random absent objects), popular (choosing most frequent objects in dataset but absent in current image), and adversarial (selecting absent objects often co-occurring with present ones).
CIEM~\cite{CIEM} is akin to POPE, yet it automates object selection by prompting ChatGPT. NOPE~\cite{NOPE} is another VQA-based method that is designed to evaluate LVLMs' ability of discerning the absence of objects in visual queries, with correct responses being negative statements.

\begin{figure*}
    \centering
    \includegraphics[width=\linewidth]{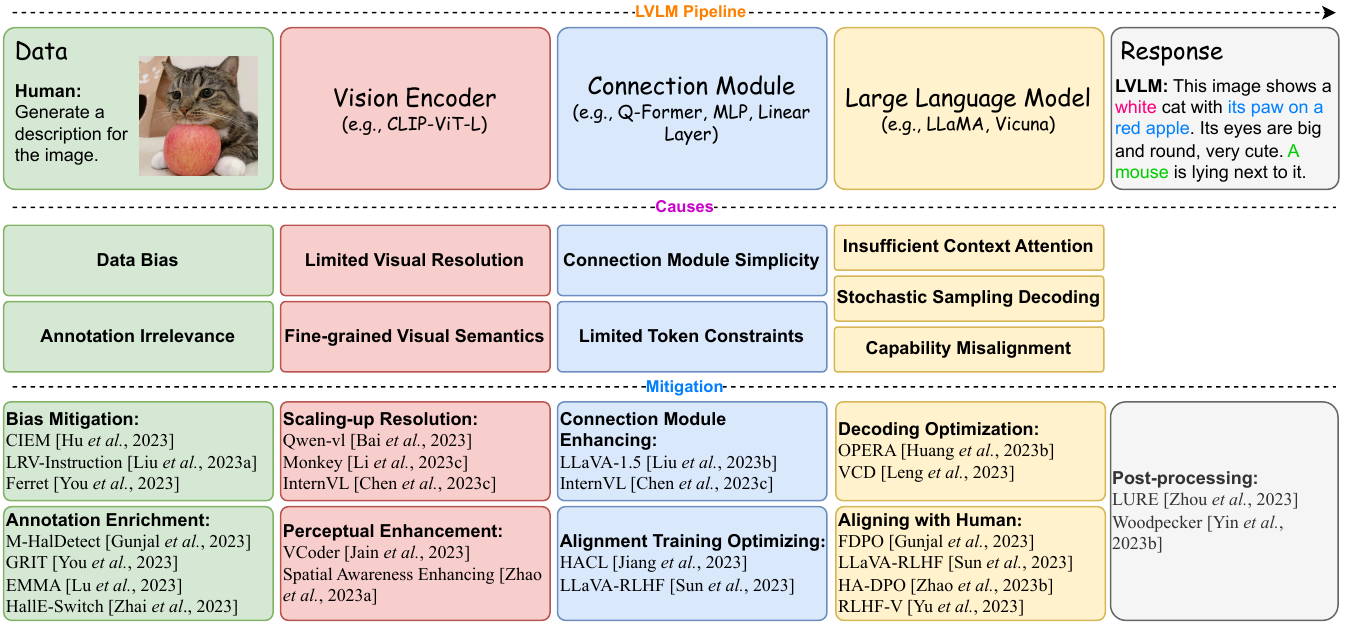}
    \caption{The causes and mitigation methods of hallucinations in LVLMs.}
    \label{fig:causes_mitigation}
\end{figure*}

\subsection{Evaluation Benchmarks}
\label{sec:evaluation_benchmarks}
Distinct from standard LVLM benchmarks that assess general LVLM capabilities, LVLM hallucination benchmarks specifically target non-hallucinatory generation or hallucination discrimination. Many such benchmarks have been proposed along with the evaluation methods. Table~\ref{tab:evaluation_benchmarks} shows a synthesis of representative benchmarks. These benchmarks are categorized based on the type of evaluation approach: Discriminative (Dis) or Generative (Gen). 

\paragraph{Discriminative Benchmarks}~POPE \cite{POPE}, NOPE \cite{NOPE}, and CIEM \cite{CIEM} are discriminative benchmarks, and their dataset sizes are 3000, 17983, and 72941, respectively. All of these three benchmarks focus only on object hallucinations and adopt accuracy as the evaluation metric, which is obtained by inquiring whether objects are present in images and comparing model responses with ground-truth answers.

\paragraph{Generative Benchmarks}~As delineated in Table~\ref{tab:evaluation_benchmarks}, current research  emphasizes generative benchmarks over discriminative ones. While discriminative benchmarks primarily evaluate hallucinations at the object level, generative benchmarks extend their scope to include a broader spectrum of hallucinations, including attribute and relation ones \cite{gunjal2023detecting,liu2023mitigating,FAITHScore,wang2023evaluation,sun2023aligning}. Notably, AMBER~\cite{wang2023llm} stands out as a comprehensive benchmark that integrates both generative and discriminative tasks. Furthermore, we observe that the metrics employed in generative benchmarks tend to be more complex and varied, in contrast to the reliance on accuracy metrics in discriminative benchmarks.
This is probably because, distinct from discriminative methods that mostly compare model outputs with ground-truth answers, generative evaluation methods require tailored metrics to target specific hallucination categories in an attempt to analyze the responses generated by LVLMs.

\section{Causes of LVLM Hallucinations}
Hallucinations in LVLMs can arise from various factors. As depicted in Figure~\ref{fig:causes_mitigation}, in correspondence with LVLM pipeline. These causes can be categorized into the following aspects:

\subsection{Hallucinations from Data}
\label{sec:causes_data}
The quality of training data largely impacts training efficiency and model performance. However, existing LVLM training data suffer from several quality challenges, fostering certain types of hallucinations.

\paragraph{Data Bias} One notable issue of data bias is the presence of \textit{distribution imbalance} in existing training data, particularly in factual judgment QA pairs where the majority of the answers are ``Yes''~\cite{CIEM,liu2023mitigating}. LVLMs trained on such biased data tend to answer ``Yes'' consistently and provide explanations even for incorrect or misleading prompts. Another issue is \textit{data homogeneity}, which impedes the model's ability to understand visual information and execute instructions under various environments~\cite {you2023ferret,liu2023mitigating}. For example, MiniGPT-4 can only describe images irrespective of the specific question posed by users due to the lack of diversified instruction learning. Similarly, LLaVA struggles to accurately describe local visual relations due to the limited varieties in visual information it was trained on.

\paragraph{Annotation Irrelevance} A large amount of instruction data is primarily synthesized from image-caption and detection data using LLMs. Nevertheless, due to the unreliability of these generative models, this approach encounters a challenge of annotation relevance: the generated long instructions often contains objects, attributes, and relationships that may not correspond to the fine-grained content depicted in the images~\cite{liu2023mitigating}. Training on such data also catalyzes hallucinations.

\subsection{Hallucinations from Vision Encoder}
\label{sec:causes_vision_encoder}

The widely adopted vision encoders in LVLMs are derived from CLIP, which maps visual and textual features to the same space by contrastive learning. Though CLIP achieves excellent performance on various visual understanding tasks, it is nonetheless limited in fully expressing visual information, such as restricted visual resolution and fine-grained visual semantics.
\paragraph{Limited Visual Resolution} Higher image resolution can make visual encoder more accurate in object recognition and perceive more visual details, thus alleviating hallucinations~\cite{HallE-Switch,li2023monkey}. However, handling a wider range of image resolutions is computation-wise demanding, hence existing models usually use smaller image resolution, such as $224\times224$ and $336\times336$ pixels for LLaVA and LLaVA-1.5 respectively.
\paragraph{Fine-grained Visual Semantics} CLIP aligns visual content and word embeddings through contrastive learning that mainly focuses only on salient objects, failing to capture fine-grained aspects of an image and resulting in hallucinations~\cite{jain2023vcoder,cho2022fine}, such as background description, object counting, and object relation.  

\subsection{Hallucinations from Modality Aligning}
\label{sec:causes_connection_module}

The connection module projects visual features into the LLM's word embedding space, aligning visual and textual modalities. Consequently, misalignment could be a key factor in hallucination generation \cite{sun2023aligning,zhao2023beyond}. Furthermore, \cite{jiang2023hallucination} reveal that even in advanced LVLMs like MiniGPT-4 and LLaVA, the gap between visual and textual features stands significant and lead to hallucinations. The simplicity of connection module and constraint of tokens are two main causes.

\paragraph{Connection Module Simplicity} Simple structures, such as linear layers, are commonly employed as connection modules to align visual and textual modalities. Although cost-efficient, the simplicity of the connection modules hinder comprehensive multimodal connection~\cite{LLaVA-1.5}, thereby increasing the risk of hallucinations.

\paragraph{Limited Token Constraints} Q-Former is a widely used multimodal alignment module in LVLMs such as BLIP-2, InstructBLIP and MiniGPT-4. It is trained with various objectives to encode a predetermined number of random initialized tokens (e.g., 32) into visual features that are aligned with text. However, the restricted quantity of these tokens prevents encoding all the information present in images during the alignment process~\cite{yin2023survey,chen2023internvl}, and consequent information loss raises the risk of hallucinations.

\subsection{Hallucinations from LLM}
\label{sec:causes_llm}

LLMs are pivotal in the architecture of LVLMs, as LLMs significantly enhance LVLMs ability to process complex multimodal tasks. However, this integration also introduce the inherent hallucination challenges of LLMs to LVLMs. A growing body of literature has explored various factors contributing to hallucinations in LLMs \cite{ji2023survey,zhang2023siren,huang2023survey}. In this section, we briefly summarize potential causes of hallucinations originating from LLMs within the scope of vision-language.

\paragraph{Insufficient Context Attention} Insufficient context attention is the phenomenon where the model, during the decoding process, focuses only on partial information from the context. The phenomenon includes focusing excessively on the current segment of generated content while ignoring input visual information~\cite{wang2023vigc,lee2023volcano}, prioritizing language patterns to produce fluent yet inaccurate content~\cite{wang2023evaluation}, and concentrating on partial summarizing tokens of the generated content, resulting in fabrication and hallucinatory content~\cite{huang2023opera}.

\paragraph{Stochastic Sampling Decoding} Stochastic sampling introduces randomness into the decoding process and is a commonly used decoding strategy in LLMs. It helps prevent the generation of low-quality text often associated with sequences of high likelihood and serves to enrich the generated content~\cite{holtzman2019curious}. However, the incorporation of randomness in the model also amplify the risk of hallucinations~\cite{chuang2023dola,dziri2021neural}.

\paragraph{Capability Misalignment} Capability misalignment in LLMs is the disparity between the model's inherent capabilities that established during the pre-training phase, and the expanded requirements imposed during the instruction tuning. This misalignment leads LLMs to produce responses beyond their established knowledge limits, increasing the potential for hallucinatory content~\cite{huang2023survey}.

\section{Mitigation of LVLM Hallucination}

Various methods have been proposed to mitigate hallucinations in LVLMs. As depicted in Figure~\ref{fig:causes_mitigation}, these methods are tailored to address the identified causes. In this section, we elaborate on them from the following perspectives:

\subsection{Mitigation for Data}

Optimizing training data is a direct and effective way to alleviate hallucinations. Regarding the explored deficiency in current datasets~(\S\ref{sec:causes_data}), some measures are taken to address the issues in the following two perspectives:
\paragraph{Bias Mitigation}
Considering the imbalance of positive and negative samples in the training data, CIEM~\cite{CIEM} utilizes off-shelf LLMs to generate contrastive question-answer pairs from annotated image-text datasets. These pairs are then leveraged in introduced Contrastive Instruction Tuning (CIT). LRV-Instruction~\cite{liu2023mitigating} proposes a large and diverse dataset consisting of 400,000 visual instructions. This dataset not only includes positive visual instructions but also incorporates negative ones on three different semantic levels. Furthermore, delving into a fine-grained manner, Ferret~\cite{you2023ferret} proposes to mine 95,000 negative samples by replacing the original categories, attributes, or quantity information with analogous fake ones. This method effectively enhances the model robustness.

\paragraph{Annotation Enrichment}
Constructing richly annotated datasets helps mitigating hallucination by supervising LVLMs to extract visual content accurately and align modalities comprehensively.
One notable dataset is M-HalDetect \cite{gunjal2023detecting}. It consists of 4,000 image-description pairs sourced from COCO labelled with object existence, relative positions, and attributes.
Another valuable dataset is GRIT \cite{you2023ferret}. GRIT features 1.1 million samples with refer-and-ground instructions over hierarchical spatial knowledge.
In addition, \cite{HallE-Switch} develops two datasets: one with RAM-system~\cite{zhang2023recognize} grounded objects and the other with objects grounded by RAM or generated by LLMs.
Lastly, \cite{lu2023evaluation} generated 180,000 fine-grained instruction samples, covering multi-round conversations, vision-prompted recognition, and fact-checking.

\subsection{Mitigation for Vision Encoder}
\paragraph{Scaling-up Vision Resolution} \cite{bai2023qwen} have shown the effectiveness of gradual enlargement of image resolution from $224\times224$ to $448\times448$. MONKEY \cite{li2023monkey} processes high-resolution images by dividing them into patches with the size matched the vision encoder, and local features extracted from patches are processed by the LLM. \cite{chen2023internvl} proposes InternVL that scales up the vision encoder to 6 billion parameters and can process images with widths ranging from 1,664 to 6,144 pixels. However, this approach is resource-demanding for pretraining with large-scale data.

\paragraph{Perceptual Enhancement} Most existing LVLMs employ the ViT from CLIP as the vision encoder, which mainly focuses only on salient objects and inevitably omit some visual cues. To enhance the object-level perceptual capability, \cite{jain2023vcoder} proposes VCoder that uses extra perception modalities, such as segmentation map, depth map, or both, as the control inputs through additional vision encoders. To enhance the spatial awareness capability, \cite{zhao2023enhancing} suggests introducing addition pre-trained models to acquire spatial position information and scene graph details, which are then used to guide the LVLMs in addressing user queries.

\subsection{Mitigation for Connection Module}

\paragraph{Connection Module Enhancing} As discussed in \S \ref{sec:causes_connection_module}, the limited capacity of the connection module hinders perfect modality alignment, increasing hallucination risks. To better align the vision and language modalities, researchers have developed more capable connection modules recently. For instance, LLaVA-1.5 enhances the connection module in LLaVA by upgrading from a single linear layer to a MLP. Moreover, \cite{chen2023internvl} utilizes LLaMA2 to construct QLLaMA, which significantly outperforms the Q-Former in aligning visual features with text.

\paragraph{Alignment Training Optimizing} Strengthening the alignment training process is an effective method to enhance modality alignment and reduce the risk of hallucinations. \cite{jiang2023hallucination} observe a significant gap between visual and textual tokens in LVLMs. By explicitly adding new learning objectives to bring visual and textual tokens closer, the hallucinations can be reduced. On the other hand, \cite{sun2023aligning} employ Reinforcement Learning from Human Feedback (RLHF)~\cite{RLHF} to align different modalities, achieving hallucination reduction.

\subsection{Mitigation for LLM}
\label{sec:mitigation_llm}

\paragraph{Decoding Optimization} In response to insufficient context attention, some recent research attempts to enable the model to focus on proper contexts during decoding, thereby mitigating hallucinations. \cite{huang2023opera} notice LVLMs may generate hallucinations by over-focusing on few summary tokens and neglecting image tokens. To mitigate hallucinations, they develop a novel decoding strategy named OPERA that modifies the beam search process with a weighted scoring system to de-prioritize over-trusted candidates and a retrospection mechanism for re-evaluating decision points. \cite{leng2023mitigating} focused on visual uncertainty and its impact on hallucinations caused by statistical bias and language priors. They proposed a visual contrastive decoding strategy, which contrasts outputs from original and altered visuals, helping correct the model's over-reliance on language priors and statistical biases.

\paragraph{Aligning with Human} Training LVLMs to align with human preferences can improve the quality of model responses. Recently, some efforts focus on preference-aligned training, where non-hallucinatory responses are favored, while hallucinatory contents are not preferred. LLaVA-RLHF~\cite{sun2023aligning} pioneers the use of RLHF for training LVLMs to align with human preferences. The authors first train a reward model to mirror human preferences. Subsequently, they apply a fact augmented RLHF and the reward model to align the LVLMs with human preferences. Another category of methods~\cite{gunjal2023detecting,zhao2023beyond,yu2023rlhf} are based on Direct Preference Optimization (DPO)~\cite{rafailov2023direct} that trains LLMs directly from human preference data, avoiding the complexities of reward-model-based reinforcement learning methods like RLHF. After constructing hallucinatory labelled training data, these work present DPO variants, notably HA-DPO~\cite{zhao2023beyond} and FDPO~\cite{gunjal2023detecting}, to train the LVLMs better align with human and generate non-hallucinatory responses.

\subsection{Mitigation via Post-processing}\label{mitigation:post-process}

Beyond manipulating modules of LVLMs, hallucinations can be mitigated through post-processing or output editing via an additional module or operations. The post-processing methods input visual data, user instructions, and LVLM responses, and outputs hallucination-free, refined responses.

More recently, two post-processing methods were introduced to reduce hallucinations in generated image descriptions: LURE~\cite{zhou2023analyzing} and Woodpecker~\cite{yin2023woodpecker}. LURE leverages insights into the causative factors of object hallucination to train a revisor model based on MiniGPT-4. Woodpecker is a training-free post-processing method comprising five stages that extract key concepts, formulate questions, apply visual knowledge validation, generate a structured visual knowledge base, and refine the generated content by correcting inaccuracies and integrating evidence from the knowledge base.

\section{Future Directions}
The research on hallucinations in LVLMs is still in its early stage, leaving considerable room for development and improvement. We detail these areas of further research below:

\paragraph{Supervision Objective} Meticulous supervision objectives are crucial for models to acquire desired capabilities or suppress unsought symptoms. Current LVLMs predominantly employ coarse-grained (e.g., image-caption)~\cite{BLIP2} alignments and whole-image instruction tuning, which limits their ability to understand complex visual details. More detailed, spatially-aware supervision such as object masked language modeling~\cite{dai2023plausible} for alignment and position-enhanced instruction following~\cite{Chen2023PVIT} hold immense potentials for further advancements.

\paragraph{Enriching Modalities} Integrating multi-modal signals not only expands the capabilities of models but also notably enhances their performance on single-modal tasks~\cite{Girdhar2023ImageBind}. This can be attributed to high-quality representation learning from correlated and complementary information.
Recent LVLMs research made strides in incorporating more modalities (e.g., videos) through methods including through cache retrieval~\cite{Han2023ImageBindLLM} and pre-alignment~\cite{Lin2023VideoLLaVA}. These studies show benefits in addressing the ``existence'' issue. Moving forward, exploration into designing advanced paradigms for coupling multiple modalities is a worthwhile avenue of investigation.

\paragraph{LVLM as Agent}
Empowered by LLMs, the understanding and reasoning capabilities of LVLMs are propelled to unprecedented levels. Yet, accurately perceiving and expressing specific visual details (like numbers in tables) is still challenging. Rather than using more advanced visual encoders with extensive training, a practical alternative is to teach LVLMs to utilize visual tools (e.g., detection and segmentation models) and reason with accurate and structured outputs. Recent research on ``LVLMs as agent'' ~\cite{Liu2023LLaVAPlus} demonstrates impressive fidelity in tasks like OCR-VQA, object tagging, and grounding, validating the viability of this direction.

\paragraph{Delving into Interpretability}
The post-processing methods~(\S\ref{mitigation:post-process}) address hallucinations in LVLMs by refining generated outputs, but require careful design and higher computational costs. Further study is needed on LVLMs' internal hallucination mechanisms and potential solutions. Several decoding strategies~(\S\ref{sec:mitigation_llm}) and model editing techniques~\cite{Cheng23MMEidt} show good reliability and effectiveness against hallucination. Future in-depth research into LVLMs' defects could lead to promising interpretability-oriented solutions.

\section{Conclusion}
Equipped with advanced vision encoders, robust LLMs, and modality connection modules, LVLMs excel in open-domain vision-language tasks. However, hallucinations significantly challenge LVLMs' practical application. In this survey, we conduct a meticulous investigation into the phenomenon of hallucinations in LVLMs. This exploration introduces innovative evaluation methods along with pertinent benchmarks, provides a detailed analysis of the fundamental causes behind these hallucinations, and discusses effective mitigation approaches. We also delve into the existing challenges and discuss potential directions. This survey aims to lay a foundation for tackle complexities of hallucinations in LVLMs and facilitate future research towards practical implementation of these models in various applications.

\bibliographystyle{named}
\bibliography{ijcai23}

\begin{thebibliography}{}

\bibitem[\protect\citeauthoryear{Alayrac \bgroup \em et al.\egroup }{2022}]{Flamingo}
Jean-Baptiste Alayrac, Jeff Donahue, Pauline Luc, et~al.
\newblock Flamingo: a visual language model for few-shot learning.
\newblock In {\em NeurIPS}, volume~35, 2022.

\bibitem[\protect\citeauthoryear{Bai \bgroup \em et al.\egroup }{2023}]{bai2023qwen}
Jinze Bai, Shuai Bai, Shusheng Yang, et~al.
\newblock Qwen-vl: A frontier large vision-language model with versatile abilities.
\newblock {\em arXiv preprint arXiv:2308.12966}, 2023.

\bibitem[\protect\citeauthoryear{Chen \bgroup \em et al.\egroup }{2023a}]{Chen2023PVIT}
Chi Chen, Ruoyu Qin, Fuwen Luo, et~al.
\newblock Position-enhanced visual instruction tuning for multimodal large language models.
\newblock {\em arXiv preprint arXiv:2308.13437}, 2023.

\bibitem[\protect\citeauthoryear{Chen \bgroup \em et al.\egroup }{2023b}]{chen2023minigpt}
Jun Chen, Deyao Zhu, Xiaoqian Shen, et~al.
\newblock Minigpt-v2: large language model as a unified interface for vision-language multi-task learning.
\newblock {\em arXiv preprint arXiv:2310.09478}, 2023.

\bibitem[\protect\citeauthoryear{Chen \bgroup \em et al.\egroup }{2023c}]{chen2023internvl}
Zhe Chen, Jiannan Wu, Wenhai Wang, et~al.
\newblock Internvl: Scaling up vision foundation models and aligning for generic visual-linguistic tasks.
\newblock {\em arXiv preprint arXiv:2312.14238}, 2023.

\bibitem[\protect\citeauthoryear{Cheng \bgroup \em et al.\egroup }{2023}]{Cheng23MMEidt}
Siyuan Cheng, Bozhong Tian, Qingbin Liu, et~al.
\newblock Can we edit multimodal large language models?
\newblock In {\em {EMNLP}}, 2023.

\bibitem[\protect\citeauthoryear{Cho \bgroup \em et al.\egroup }{2022}]{cho2022fine}
Jaemin Cho, Seunghyun Yoon, Ajinkya Kale, et~al.
\newblock Fine-grained image captioning with clip reward.
\newblock In {\em Findings of NAACL}, 2022.

\bibitem[\protect\citeauthoryear{Chuang \bgroup \em et al.\egroup }{2023}]{chuang2023dola}
Yung-Sung Chuang, Yujia Xie, Hongyin Luo, et~al.
\newblock Dola: Decoding by contrasting layers improves factuality in large language models.
\newblock {\em arXiv preprint arXiv:2309.03883}, 2023.

\bibitem[\protect\citeauthoryear{Dai \bgroup \em et al.\egroup }{2023a}]{InstructBLIP}
Wenliang Dai, Junnan Li, Dongxu Li, et~al.
\newblock Instructblip: Towards general-purpose vision-language models with instruction tuning.
\newblock {\em arXiv preprint arXiv:2305.06500}, 2023.

\bibitem[\protect\citeauthoryear{Dai \bgroup \em et al.\egroup }{2023b}]{dai2023plausible}
Wenliang Dai, Zihan Liu, Ziwei Ji, et~al.
\newblock Plausible may not be faithful: Probing object hallucination in vision-language pre-training.
\newblock In {\em EACL}, 2023.

\bibitem[\protect\citeauthoryear{Dziri \bgroup \em et al.\egroup }{2021}]{dziri2021neural}
Nouha Dziri, Andrea Madotto, Osmar~R Zaiane, et~al.
\newblock Neural path hunter: Reducing hallucination in dialogue systems via path grounding.
\newblock In {\em EMNLP}, 2021.

\bibitem[\protect\citeauthoryear{Gao \bgroup \em et al.\egroup }{2023}]{gao2023llama}
Peng Gao, Jiaming Han, Renrui Zhang, et~al.
\newblock Llama-adapter v2: Parameter-efficient visual instruction model.
\newblock {\em arXiv preprint arXiv:2304.15010}, 2023.

\bibitem[\protect\citeauthoryear{Girdhar \bgroup \em et al.\egroup }{2023}]{Girdhar2023ImageBind}
Rohit Girdhar, Alaaeldin El{-}Nouby, Zhuang Liu, et~al.
\newblock Imagebind one embedding space to bind them all.
\newblock In {\em {CVPR}}, 2023.

\bibitem[\protect\citeauthoryear{Gunjal \bgroup \em et al.\egroup }{2023}]{gunjal2023detecting}
Anisha Gunjal, Jihan Yin, and Erhan Bas.
\newblock Detecting and preventing hallucinations in large vision language models.
\newblock {\em arXiv preprint arXiv:2308.06394}, 2023.

\bibitem[\protect\citeauthoryear{Han \bgroup \em et al.\egroup }{2023}]{Han2023ImageBindLLM}
Jiaming Han, Renrui Zhang, Wenqi Shao, et~al.
\newblock Imagebind-llm: Multi-modality instruction tuning.
\newblock {\em arXiv preprint arXiv:2309.03905}, 2023.

\bibitem[\protect\citeauthoryear{Holtzman \bgroup \em et al.\egroup }{2020}]{holtzman2019curious}
Ari Holtzman, Jan Buys, Li~Du, et~al.
\newblock The curious case of neural text degeneration.
\newblock In {\em ICLR}, 2020.

\bibitem[\protect\citeauthoryear{Hu \bgroup \em et al.\egroup }{2022}]{LoRA}
Edward~J. Hu, Yelong Shen, Phillip Wallis, et~al.
\newblock Lora: Low-rank adaptation of large language models.
\newblock In {\em ICLR}, 2022.

\bibitem[\protect\citeauthoryear{Hu \bgroup \em et al.\egroup }{2023}]{CIEM}
Hongyu Hu, Jiyuan Zhang, Minyi Zhao, et~al.
\newblock Ciem: Contrastive instruction evaluation method for better instruction tuning.
\newblock In {\em NeurIPS 2023 Workshop on Instruction Tuning and Instruction Following}, 2023.

\bibitem[\protect\citeauthoryear{Huang \bgroup \em et al.\egroup }{2023a}]{huang2023survey}
Lei Huang, Weijiang Yu, Weitao Ma, et~al.
\newblock A survey on hallucination in large language models: Principles, taxonomy, challenges, and open questions.
\newblock {\em arXiv preprint arXiv:2311.05232}, 2023.

\bibitem[\protect\citeauthoryear{Huang \bgroup \em et al.\egroup }{2023b}]{huang2023opera}
Qidong Huang, Xiaoyi Dong, Pan Zhang, et~al.
\newblock Opera: Alleviating hallucination in multi-modal large language models via over-trust penalty and retrospection-allocation.
\newblock {\em arXiv preprint arXiv:2311.17911}, 2023.

\bibitem[\protect\citeauthoryear{Jain \bgroup \em et al.\egroup }{2023}]{jain2023vcoder}
Jitesh Jain, Jianwei Yang, and Humphrey Shi.
\newblock Vcoder: Versatile vision encoders for multimodal large language models.
\newblock {\em arXiv preprint arXiv:2312.14233}, 2023.

\bibitem[\protect\citeauthoryear{Ji \bgroup \em et al.\egroup }{2023}]{ji2023survey}
Ziwei Ji, Nayeon Lee, Rita Frieske, et~al.
\newblock Survey of hallucination in natural language generation.
\newblock {\em ACM Computing Surveys}, 55(12), 2023.

\bibitem[\protect\citeauthoryear{Jiang \bgroup \em et al.\egroup }{2023}]{jiang2023hallucination}
Chaoya Jiang, Haiyang Xu, Mengfan Dong, et~al.
\newblock Hallucination augmented contrastive learning for multimodal large language model.
\newblock {\em arXiv preprint arXiv:2312.06968}, 2023.

\bibitem[\protect\citeauthoryear{Jing \bgroup \em et al.\egroup }{2023}]{FAITHScore}
Liqiang Jing, Ruosen Li, Yunmo Chen, et~al.
\newblock Faithscore: Evaluating hallucinations in large vision-language models.
\newblock {\em arXiv preprint arXiv:2311.01477}, 2023.

\bibitem[\protect\citeauthoryear{Lee \bgroup \em et al.\egroup }{2023}]{lee2023volcano}
Seongyun Lee, Sue~Hyun Park, Yongrae Jo, et~al.
\newblock Volcano: Mitigating multimodal hallucination through self-feedback guided revision.
\newblock {\em arXiv preprint arXiv:2311.07362}, 2023.

\bibitem[\protect\citeauthoryear{Leng \bgroup \em et al.\egroup }{2023}]{leng2023mitigating}
Sicong Leng, Hang Zhang, Guanzheng Chen, et~al.
\newblock Mitigating object hallucinations in large vision-language models through visual contrastive decoding.
\newblock {\em arXiv preprint arXiv:2311.16922}, 2023.

\bibitem[\protect\citeauthoryear{Li \bgroup \em et al.\egroup }{2023a}]{BLIP2}
Junnan Li, Dongxu Li, Silvio Savarese, et~al.
\newblock Blip-2: Bootstrapping language-image pre-training with frozen image encoders and large language models.
\newblock In {\em ICML}, 2023.

\bibitem[\protect\citeauthoryear{Li \bgroup \em et al.\egroup }{2023b}]{POPE}
Yifan Li, Yifan Du, Kun Zhou, et~al.
\newblock Evaluating object hallucination in large vision-language models.
\newblock In {\em EMNLP}, 2023.

\bibitem[\protect\citeauthoryear{Li \bgroup \em et al.\egroup }{2023c}]{li2023monkey}
Zhang Li, Biao Yang, Qiang Liu, et~al.
\newblock Monkey: Image resolution and text label are important things for large multi-modal models.
\newblock {\em arXiv preprint arXiv:2311.06607}, 2023.

\bibitem[\protect\citeauthoryear{Lin \bgroup \em et al.\egroup }{2023}]{Lin2023VideoLLaVA}
Bin Lin, Bin Zhu, Yang Ye, et~al.
\newblock Video-llava: Learning united visual representation by alignment before projection.
\newblock {\em arXiv preprint arXiv:2311.10122}, 2023.

\bibitem[\protect\citeauthoryear{Liu \bgroup \em et al.\egroup }{2023a}]{liu2023mitigating}
Fuxiao Liu, Kevin Lin, Linjie Li, et~al.
\newblock Mitigating hallucination in large multi-modal models via robust instruction tuning.
\newblock {\em arXiv preprint arXiv:2306.14565}, 2023.

\bibitem[\protect\citeauthoryear{Liu \bgroup \em et al.\egroup }{2023b}]{LLaVA-1.5}
Haotian Liu, Chunyuan Li, Yuheng Li, et~al.
\newblock Improved baselines with visual instruction tuning.
\newblock {\em arXiv preprint arXiv:2310.03744}, 2023.

\bibitem[\protect\citeauthoryear{Liu \bgroup \em et al.\egroup }{2023c}]{LLaVA}
Haotian Liu, Chunyuan Li, Qingyang Wu, et~al.
\newblock Visual instruction tuning.
\newblock In {\em NeurIPS}, 2023.

\bibitem[\protect\citeauthoryear{Liu \bgroup \em et al.\egroup }{2023d}]{Liu2023LLaVAPlus}
Shilong Liu, Hao Cheng, Haotian Liu, et~al.
\newblock Llava-plus: Learning to use tools for creating multimodal agents.
\newblock {\em arXiv preprint arXiv:2311.05437}, 2023.

\bibitem[\protect\citeauthoryear{Long \bgroup \em et al.\egroup }{2022}]{long2022vision}
Siqu Long, Feiqi Cao, Soyeon~Caren Han, et~al.
\newblock Vision-and-language pretrained models: A survey.
\newblock In {\em IJCAI}, 2022.

\bibitem[\protect\citeauthoryear{Lovenia \bgroup \em et al.\egroup }{2023}]{NOPE}
Holy Lovenia, Wenliang Dai, Samuel Cahyawijaya, et~al.
\newblock Negative object presence evaluation (nope) to measure object hallucination in vision-language models.
\newblock {\em arXiv preprint arXiv:2310.05338}, 2023.

\bibitem[\protect\citeauthoryear{Lu \bgroup \em et al.\egroup }{2018}]{lu2018neural}
Jiasen Lu, Jianwei Yang, Dhruv Batra, et~al.
\newblock Neural baby talk.
\newblock In {\em CVPR}, 2018.

\bibitem[\protect\citeauthoryear{Lu \bgroup \em et al.\egroup }{2023}]{lu2023evaluation}
Jiaying Lu, Jinmeng Rao, Kezhen Chen, et~al.
\newblock Evaluation and mitigation of agnosia in multimodal large language models.
\newblock {\em arXiv preprint arXiv:2309.04041}, 2023.

\bibitem[\protect\citeauthoryear{OpenAI}{2023}]{GPT-4}
OpenAI.
\newblock Gpt-4 technical report.
\newblock {\em arXiv preprint arXiv:2303.08774}, 2023.

\bibitem[\protect\citeauthoryear{Radford \bgroup \em et al.\egroup }{2021}]{CLIP}
Alec Radford, Jong~Wook Kim, Chris Hallacy, et~al.
\newblock Learning transferable visual models from natural language supervision.
\newblock In {\em ICML}, 2021.

\bibitem[\protect\citeauthoryear{Rafailov \bgroup \em et al.\egroup }{2023}]{rafailov2023direct}
Rafael Rafailov, Archit Sharma, Eric Mitchell, et~al.
\newblock Direct preference optimization: Your language model is secretly a reward model.
\newblock {\em arXiv preprint arXiv:2305.18290}, 2023.

\bibitem[\protect\citeauthoryear{Rohrbach \bgroup \em et al.\egroup }{2018}]{CHAIR}
Anna Rohrbach, Lisa~Anne Hendricks, Kaylee Burns, et~al.
\newblock Object hallucination in image captioning.
\newblock In {\em EMNLP}, 2018.

\bibitem[\protect\citeauthoryear{Stiennon \bgroup \em et al.\egroup }{2020}]{RLHF}
Nisan Stiennon, Long Ouyang, Jeffrey Wu, et~al.
\newblock Learning to summarize with human feedback.
\newblock In {\em NeurIPS}, volume~33, 2020.

\bibitem[\protect\citeauthoryear{Sun \bgroup \em et al.\egroup }{2023}]{sun2023aligning}
Zhiqing Sun, Sheng Shen, Shengcao Cao, et~al.
\newblock Aligning large multimodal models with factually augmented rlhf.
\newblock {\em arXiv preprint arXiv:2309.14525}, 2023.

\bibitem[\protect\citeauthoryear{Touvron \bgroup \em et al.\egroup }{2023a}]{LLaMA}
Hugo Touvron, Thibaut Lavril, Gautier Izacard, et~al.
\newblock Llama: Open and efficient foundation language models.
\newblock {\em arXiv preprint arXiv:2302.13971}, 2023.

\bibitem[\protect\citeauthoryear{Touvron \bgroup \em et al.\egroup }{2023b}]{touvron2023llama}
Hugo Touvron, Louis Martin, Kevin Stone, et~al.
\newblock Llama 2: Open foundation and fine-tuned chat models.
\newblock {\em arXiv preprint arXiv:2307.09288}, 2023.

\bibitem[\protect\citeauthoryear{Wang \bgroup \em et al.\egroup }{2023a}]{wang2023vigc}
Bin Wang, Fan Wu, Xiao Han, et~al.
\newblock Vigc: Visual instruction generation and correction.
\newblock {\em arXiv preprint arXiv:2308.12714}, 2023.

\bibitem[\protect\citeauthoryear{Wang \bgroup \em et al.\egroup }{2023b}]{wang2023llm}
Junyang Wang, Yuhang Wang, Guohai Xu, et~al.
\newblock An llm-free multi-dimensional benchmark for mllms hallucination evaluation.
\newblock {\em arXiv preprint arXiv:2311.07397}, 2023.

\bibitem[\protect\citeauthoryear{Wang \bgroup \em et al.\egroup }{2023c}]{wang2023evaluation}
Junyang Wang, Yiyang Zhou, Guohai Xu, et~al.
\newblock Evaluation and analysis of hallucination in large vision-language models.
\newblock {\em arXiv preprint arXiv:2308.15126}, 2023.

\bibitem[\protect\citeauthoryear{Yin \bgroup \em et al.\egroup }{2023a}]{yin2023survey}
Shukang Yin, Chaoyou Fu, Sirui Zhao, et~al.
\newblock A survey on multimodal large language models.
\newblock {\em arXiv preprint arXiv:2306.13549}, 2023.

\bibitem[\protect\citeauthoryear{Yin \bgroup \em et al.\egroup }{2023b}]{yin2023woodpecker}
Shukang Yin, Chaoyou Fu, Sirui Zhao, et~al.
\newblock Woodpecker: Hallucination correction for multimodal large language models.
\newblock {\em arXiv preprint arXiv:2310.16045}, 2023.

\bibitem[\protect\citeauthoryear{You \bgroup \em et al.\egroup }{2023}]{you2023ferret}
Haoxuan You, Haotian Zhang, Zhe Gan, et~al.
\newblock Ferret: Refer and ground anything anywhere at any granularity.
\newblock {\em arXiv preprint arXiv:2310.07704}, 2023.

\bibitem[\protect\citeauthoryear{Yu \bgroup \em et al.\egroup }{2023}]{yu2023rlhf}
Tianyu Yu, Yuan Yao, Haoye Zhang, et~al.
\newblock Rlhf-v: Towards trustworthy mllms via behavior alignment from fine-grained correctional human feedback.
\newblock {\em arXiv preprint arXiv:2312.00849}, 2023.

\bibitem[\protect\citeauthoryear{Zhai \bgroup \em et al.\egroup }{2023}]{HallE-Switch}
Bohan Zhai, Shijia Yang, Xiangchen Zhao, et~al.
\newblock Halle-switch: Rethinking and controlling object existence hallucinations in large vision language models for detailed caption.
\newblock {\em arXiv preprint arXiv:2310.01779}, 2023.

\bibitem[\protect\citeauthoryear{Zhang \bgroup \em et al.\egroup }{2023a}]{zhang2023recognize}
Youcai Zhang, Xinyu Huang, Jinyu Ma, et~al.
\newblock Recognize anything: A strong image tagging model.
\newblock {\em arXiv preprint arXiv:2306.03514}, 2023.

\bibitem[\protect\citeauthoryear{Zhang \bgroup \em et al.\egroup }{2023b}]{zhang2023siren}
Yue Zhang, Yafu Li, Leyang Cui, et~al.
\newblock Siren's song in the ai ocean: A survey on hallucination in large language models.
\newblock {\em arXiv preprint arXiv:2309.01219}, 2023.

\bibitem[\protect\citeauthoryear{Zhao \bgroup \em et al.\egroup }{2023a}]{zhao2023enhancing}
Yongqiang Zhao, Zhenyu Li, Zhi Jin, et~al.
\newblock Enhancing the spatial awareness capability of multi-modal large language model.
\newblock {\em arXiv preprint arXiv:2310.20357}, 2023.

\bibitem[\protect\citeauthoryear{Zhao \bgroup \em et al.\egroup }{2023b}]{zhao2023beyond}
Zhiyuan Zhao, Bin Wang, Linke Ouyang, et~al.
\newblock Beyond hallucinations: Enhancing lvlms through hallucination-aware direct preference optimization.
\newblock {\em arXiv preprint arXiv:2311.16839}, 2023.

\bibitem[\protect\citeauthoryear{Zhou \bgroup \em et al.\egroup }{2023}]{zhou2023analyzing}
Yiyang Zhou, Chenhang Cui, Jaehong Yoon, et~al.
\newblock Analyzing and mitigating object hallucination in large vision-language models.
\newblock In {\em NeurIPS 2023 Workshop on Instruction Tuning and Instruction Following}, 2023.

\bibitem[\protect\citeauthoryear{Zhu \bgroup \em et al.\egroup }{2023}]{MiniGPT-4}
Deyao Zhu, Jun Chen, Xiaoqian Shen, et~al.
\newblock Minigpt-4: Enhancing vision-language understanding with advanced large language models.
\newblock {\em arXiv preprint arXiv:2304.10592}, 2023.

\end{thebibliography}

\end{document}